# Countering Catastrophic Forgetting of Large Language Models for Better Instruction Following via Weight-Space Model Merging

Mengxian Lyu, MS[1], Cheng Peng, PhD[1], Ziyi Chen, MS[1], Mengyuan Zhang, MS[1], Jieting Li Lu, BS[2], Yonghui Wu, PhD[1,3*]

[1]Department of Health Outcomes and Biomedical Informatics, College of Medicine, University of Florida, Gainesville, FL, USA; [2]Department of Engineering Education, Herbert Wertheim College of Engineering, University of Florida, Gainesville, FL, USA; [3]Preston A. Wells, Jr. Center for Brain Tumor Therapy, Lillian S. Wells Department of Neurosurgery, University of Florida, Gainesville, Florida, USA.

**Abstract**

*Large language models have been adopted in the medical domain for clinical documentation to reduce clinician burden. However, studies have reported that LLMs often "forgetting" a significant amount of instruction-following ability when fine-tuned using a task-specific medical dataset, a critical challenge in adopting general-purpose LLMs for clinical applications. This study presents a model merging framework to efficiently adopt general-purpose LLMs into the medical domain by countering the forgetting issue. By merging a clinical foundation model (GatorTronLlama) with a general instruct model (Llama-3.1-8B-Instruct) via interpolation merge methods, we seek to derive a domain-adapted model with good performance in clinical tasks while retaining the instruction following ability. Comprehensive evaluation across medical benchmarks and five clinical generation tasks (e.g., radiology and discharge summarization) shows that merged models could effectively mitigate catastrophic forgetting, retain clinical domain expertise while reserving instruction-following. In addition, our model merging strategies demonstrate training efficiency, achieving performance on par with fully fine-tuned baselines under severely constrained supervision (e.g., 64-shot vs. 256-shot). Consequently, weight-space merging constitutes a highly scalable solution for adopting open-source LLMs for clinical applications, facilitating broader deployment in resource-constrained healthcare environments.*

## Introduction

Physicians constantly use clinical documents as a primary way of communication for continuous patient care, interdepartmental coordination, and clinical decision-making[1]. However, this has become an increasing burden contributing to physician burnout[2,3]. Large language models (LLMs) have demonstrated potential in automating these clinical workflows by generating discharge summaries[4,5], formulating patient problem lists[6], and drafting radiology reports[7]. General-purpose LLMs lack clinical context for healthcare applications; thus, many efforts focused on domain adaptation of non-medical LLMs to the clinical domain through continuous pre-training or fine-tuning using medical corpora, developing clinical foundation models like GatorTron[8], GatorTronGPT[9], Meditron[10] , and Me-Llama[11]. While these medical LLMs capture clinical context and knowledge, many studies have reported that they typically forget instruction following abilities that are typically developed before the domain adaptation, a critical issue known as catastrophic forgetting, jeopardizing the adaptation of LLMs for clinical applications.

The instruction following ability of LLMs is usually developed through post-training using Reinforcement Learning from Human Feedback[12] (RLHF) using domain-specific instruction datasets[13]. However, this post-training is highly resource-intensive, and the instruction-following datasets are often not available for such post-training. Curating high-quality, specialized clinical instruction data is expensive, and task-specific supervised fine-tuning (SFT) usually causes catastrophic forgetting of the instruction-following ability acquired during pre-training[14].

Model merging is a promising, compute-efficient method that combines the parameters of multiple distinct models into a single checkpoint without using extensive post-training[15]. By mixing the model weights of LLMs, the model merging techniques are expected to transfer knowledge, such as reasoning or instruction-following, between models. Despite promising results in general domains, model merging is underexplored in the clinical domain. It is unclear whether the instruction-following ability of general-purpose LLMs can be transferred to clinical LLMs through model merging.

In this study, we systematically evaluate model merging as a low-resource alternative to adopt general-purpose LLMs for clinical tasks. Specifically, we first adopted an open-source LLM, Llama, into the clinical domain through continuous pretraining using clinical notes at the University of Florida Health, denoted as GatorTronLlama. Then, we explored model merging algorithms to merge GatorTronLlama with a general instruct model, Llama-3.1-8B-Instruct[16] , using Linear[17] and Spherical Linear Interpolation (SLERP) strategies. Then, we map the Pareto frontier of merged checkpoints to balance clinical domain knowledge and instruction-following abilities. We comprehensively evaluate the zero-shot, few-shot, and full SFT performance of these merged models across five clinical generation tasks. Our results demonstrate that the model merging techniques could counter the catastrophic forgetting issue to bridge the gap between clinical knowledge and general instruction-following ability as an efficient solution in adopting general-purpose LLMs to the medical domain while reserving the instruction-following ability.

## Methods
### Overview and Study Design

Based on an open-source LLM, Llama, we performed continuous pretraining using 166 billion tokens of UF Health clinical text to adopt Llama into a clinical LLM, GatorTronLlama. We conducted a two-phase experimental design to evaluate model merging for clinical applications (**Figure 1**). In the first phase, we conduct a merge recipe search using Linear and Spherical Linear Interpolation (SLERP) methods to combine GatorTronLlama with a general-purpose instruct model, Llama-3.1-8B-Instruct. The resulting merged checkpoints are evaluated in a zero-shot setting using (1) a set of clinical knowledge benchmarks, and (2) a set of instruction-following benchmark datasets (IFEval). We seek to identify Pareto-optimal checkpoints that effectively balance medical knowledge of GatorTronLlama with the instruction-following ability of Llama-Instruct. In the second phase, we adapt these Pareto-optimal checkpoints for downstream clinical applications. We run post-merge supervised fine-tuning across diverse clinical generation tasks, including radiology report summarization, discharge summary generation, patient problem list creation, and dialogue-to-notes tasks. By employing both few-shot and full-parameter SFT learning strategies, we systematically assess the data efficiency and downstream robustness relative to the original base models.

**Figure 1.** Overview of the study design and experimental workflow.

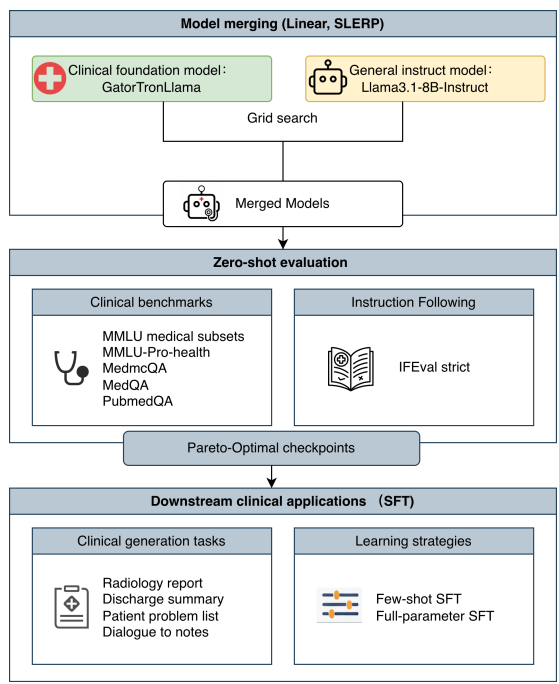

### Models

This study examines three LLMs：

**A clinical LLM (GatorTronLlama):** We utilize GatorTronLlama as a clinically oriented LLM through continuous pretraining. This model is based on the Llama-3.1-8B-Base architecture and developed through continual pre-training (CPT) on a 282-billion-token corpus comprising PubMed literature, clinical guidelines, and de-identified UF Health notes. GatorTronLlama is used as a pre-merge baseline to measure the clinical knowledge and instruction-following ability before applying model merging.

**A general instruct LLM (Llama-3.1-8B-Instruct):** We use Llama-3.1-8B-Instruct as the reference for general-purpose instruction following. Its instruction-following is optimized by Meta via Supervised Fine-Tuning (SFT) and Reinforcement Learning from Human Feedback (RLHF), serving as the instruction-expert model. We seek to examine the "forgetting" issue of domain adaptation by comparing GatorTronLlama and Llama-3.1-8B-Instruct using the instruction following dataset.

**A supervised fine-tuning LLM (GatorTronLlama_SFT):** We fine-tuned GatorTronLlama through SFT using UltraMedical[18] instruction dataset. We follow the UltraMedical instruction formatting provided by the dataset and apply the same optimization configuration as our downstream SFT tasks to ensure a fair comparison.

### Model Merging

Let $\theta_c$ denote the parameters of the clinical foundation model (GatorTronLlama) and $\theta_i$ denote the parameters of the general instruct model (Llama-3.1-8B-Instruct). We perform model merging to derive merged parameters $\theta_m$ using three representative merging strategies:

**Linear:** Computes a convex, parameter-wise weighted average of the two input model checkpoints. Specifically, given clinical parameters $\theta_c$ and instruct parameters $\theta_i$, we form $\theta_m = (1-\alpha)\theta_c + \alpha\theta_i$, which $\alpha \in [0,1]$ controls the blend ratio. This method is a simple and widely used baseline for model merging, providing an interpretable hyperparameter for balancing to trade off domain-specific performance against instruction-following behavior.

**Spherical linear interpolation (SLERP):** SLERP performs spherical interpolation between two checkpoints in weight space, producing a continuous path along a hypersphere rather than a straight line. This construction aims to preserve the overall magnitude (norm) of parameter vectors more consistently than linear blending when the two checkpoints are not well aligned.

### Merge implementation and hyperparameter Setup

We perform a grid-based merge recipe search to identify competitive hyperparameter regions for each merging family. For both Linear interpolation and SLERP, we sweep the interpolation weight $\alpha(\text{or } t) \in \{0.0, 0.1, \ldots, 1.0\}$. All merges are implemented using mergekit. Because our objective is inherently multi-criteria, we select candidate recipes by jointly considering (i) clinical benchmark performance and (ii) instruction-following performance, and we report recipes that lie on or near the Pareto frontier of these two metrics.

### Benchmarks and Evaluation Protocol

We evaluate all baseline and merged models using lm-evaluation-harness[19] (lm-eval under a consistent inference protocol. Our evaluation covers two complementary capability dimensions: clinical knowledge and instruction following.

**Clinical knowledge benchmarks:** We evaluate clinical knowledge using a set of medical knowledge benchmarks. Specifically, we use the MMLU[20] medical subsets (consisting of anatomy, clinical_knowledge, college_biology, college_medicine, medical_genetics, and professional_medicine), together with MMLU-Pro[21] health subset, to quantify clinically relevant knowledge. In addition, we include MedMCQA[22] and MedQA[23] (4-option) as domain-specific medical QA benchmarks, and PubMedQA[24] to assess biomedical question answering grounded in scientific literature. We compute Medical Avg as the unweighted mean accuracy across this clinical benchmark suite.

**Instruction-following benchmark:** We use IFEval[25] to measure instruction-following reliability, reporting the average of prompt-level strict accuracy and instance-level strict accuracy. This benchmark assesses how well merged checkpoints retain general instruction-following ability.

### Merge Recipe Selection

Because model merging introduces an inherent trade-off between clinical benchmark performance and instruction-following capability, we select merge recipes using a multi-criteria strategy rather than optimizing a single metric. We evaluate each merged checkpoint on the clinical knowledge benchmark suite and on the instruction-following benchmark (average of prompt-level and instance-level strict scores). We identify configurations that lie on or near the Pareto frontier between these two objectives. We prioritize recipes that preserve strong performance on clinical benchmarks while maintaining acceptable instruction-following performance, promoting a small set of representative merged checkpoints for downstream supervised fine-tuning and robustness analyses.

**Post-merge Supervised Fine-tuning**

To evaluate downstream clinical utility and data efficiency after merging, we conduct supervised fine-tuning (SFT) on five open-sourced clinical generation tasks:

**Brief hospital course summarization:** We use the MIMIC-IV-Ext-BHC[26] dataset, a curated collection of de-identified discharge summaries paired with their corresponding "Brief Hospital Course (BHC)" sections derived from the MIMIC-IV-Note corpus. Each instance is formatted as a supervised summarization pair, where the full discharge note serves as the input and the BHC section is the target summary.

**Radiology report summarization:** We utilize the Indiana University Chest X-Ray Collection[27] (IU X-ray) dataset, which contains chest radiograph studies paired with diagnostic reports released via the OpenI repository. Reports typically include "Findings" and "Impression" sections. We formulate summarization as Findings to Impression, treating the Findings section as the input and the Impression section as the target, reflecting the clinical workflow of converting detailed observations into a concise interpretive summary.

**Dialogue to note/note section generation:** These two datasets are derived from the MEDIQA-Chat 2023[28] shared-task. Specifically, Dialogue to note section on MTS-Dialog[29], where the model generates a single note section conditioned on a short clinician–patient conversation. ACI-Bench[30], where the model generates a full encounter note from the conversation.

**Patient active problem list generation:** For problem list generation, we use the BioNLP Workshop 2023 Shared Task 1A[31] dataset, derived from MIMIC-III[32] progress notes. Each instance is constructed from physician/resident daily progress notes in a SOAP-style format, and the target is a problem list summarizing active diagnoses/problems.

**Few-shot and SFT Experiment Settings**

**Few-shot**: We follow a few-shot learning curve protocol with shot sizes {8, 16, 32, 64, 128, 256, 512}, and additionally include the full-data setting. For each shot size, we subsample the training split of each task to the specified number of examples and fine-tune models under the same training configuration to isolate the effect of supervision budget.

**SFT:** We perform SFT with a consistent configuration across tasks and model initializations: 15 epochs, learning rate $5 \times 10^{-5}$, and effective batch size 16 (per-device batch size 4 with gradient accumulation steps 4). The maximum sequence length is set to 8192 during training; sequence truncation/packing follows the default setting of the Huggingface training framework. To avoid overly short runs in low-resource regimes, we enforce a minimum number of optimization steps: 200 steps for standard/full-data runs and 100 steps for few-shot runs.

We compare three types of initializations: (i) the clinical foundation model (GatorTronLlama), (ii) the instruction-tuned clinical baseline (GatorTronLlama + UltraMedical SFT), and (iii) the selected merged checkpoints (one best recipe per merging family: Linear, SLERP) using matched SFT settings to ensure comparability.

**Inference and Post-SFT Evaluation**

**Inference:** For downstream generation, we use batched decoding with a batch size of 16. We cap the input context at 8192 tokens and allow up to 8192 generated tokens. Unless otherwise specified, we employ stochastic sampling with temperature = 0.3 and top_p = 0.9 and apply a repetition penalty of 1.2.

**Metrics:** For downstream clinical generation tasks, we evaluate the generated results using BLEU, METEOR, ROUGE-1, ROUGE-2, ROUGE-L, and BERTScore (using F1 measures where applicable). To summarize overall trends across multiple automatic metrics, we report a composite overall score computed as the unweighted arithmetic mean of these six metric scores. Following SFT, we also re-evaluate models on the original clinical knowledge and instruction-following benchmarks to assess merged model robustness and potential catastrophic forgetting.

**Results**

To quantify the computational profile, **Table 1** compares model merging with traditional instruction fine-tuning. Constructing the Linear/SLERP model merging required 40 minutes to generate 9 merged checkpoints (merge only; excluding evaluation). In contrast, instruction fine-tuning on UltraMedical dataset required 18h 20m on a single B200 (192GB) GPU under standard SFT settings (3 epochs, learning rate 2e-5, maximum sequence length 8192). Model merging enables a training-free solution, substantially reducing optimization overhead compared to gradient-based instruction tuning.

**Table 1.** Resource comparison for model merge vs. instruction fine-tuning

| Approach | Gradient-based training | Labeled data | Wall-clock time | Hardware |
| --- | --- | --- | --- | --- |
| Model Merge | Not required | Not required | 40 min to generate 9 checkpoints | 1× B200 (192GB) |
| Instruction fine-tuning | Yes | 409,593 samples | 18h 20min | 1× B200 (192GB) |

**Figure 2** compares model performance on zero-shot settings, where the x-axis represents the instruction following dimension measured by instruction-following datasets, and the y-axis represents the clinical knowledge dimension measured by the clinical knowledge datasets. Per evaluation, GatorTronLlama demonstrates better clinical domain performance (Medical Avg: 0.6896) than Llama-3.1-8B-Instruct, but lacks instruction adherence (IFEval: 0.2244). Llama-3.1-8B-Instruct (general instruct model) shows better performance on instruction-following (IFEval: 0.5253) but shows lower clinical benchmark performance (0.6845), compared to GatorTronLlama. The merged models demonstrate a better mix of both clinical knowledge and instruction following. Several merged models outperformed both GatorTronLlama and Llama-3.1-8B-Instruct. For instance, SLERP (t=0.4) achieved a medical average of 0.7039, representing a 1.43% improvement on the clinical tasks compared with GatorTronLlama. Regarding instruction-following capability, the Llama-3.1-8B-Instruct model remains the performance upper bound (0.5253). While the merged models do not exceed this baseline, several merged models demonstrate instruction following performances close to the upper bound. SLERP (t=0.7) achieved an instruction-following score of 0.5166, recovering 98.3% of the instruction following performance while keeping a better performance on clinical tasks (0.6969 vs. 0.6845). This indicates that model merging can serve as an efficient algorithm to adopt open-source LLM into the medical domain while retaining the instruction following ability.

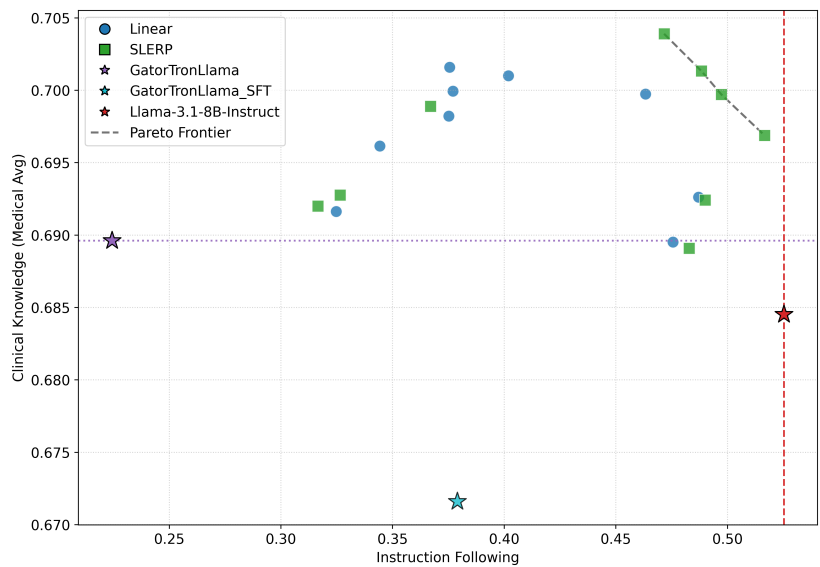

**Figure 3** shows the non-monotonic relationship between the parameter merge ratio (α or t) and model performance across both merging strategies: Linear (**Figure 3a**) and SLERP (**Figure 3b**). As shown in the trajectories, clinical-domain performance (Medical Avg) initially increases, peaking at moderate interpolation weights (around 0.4 to 0.6 depending on the strategy), before declining as the ratio shifts further toward the instruct baseline. This inflection suggests that domain expertise may benefit from a limited injection of alignment parameters, while excessive mixing could dilute specialized representations. Concurrently, instruction-following scores (IFEval) improve markedly with instruct weights but eventually plateau or slightly decline at extreme ratios (e.g., 0.8–0.9). This saturation indicates diminishing returns at high merge ratios, approaching the instruction-tuned regime.

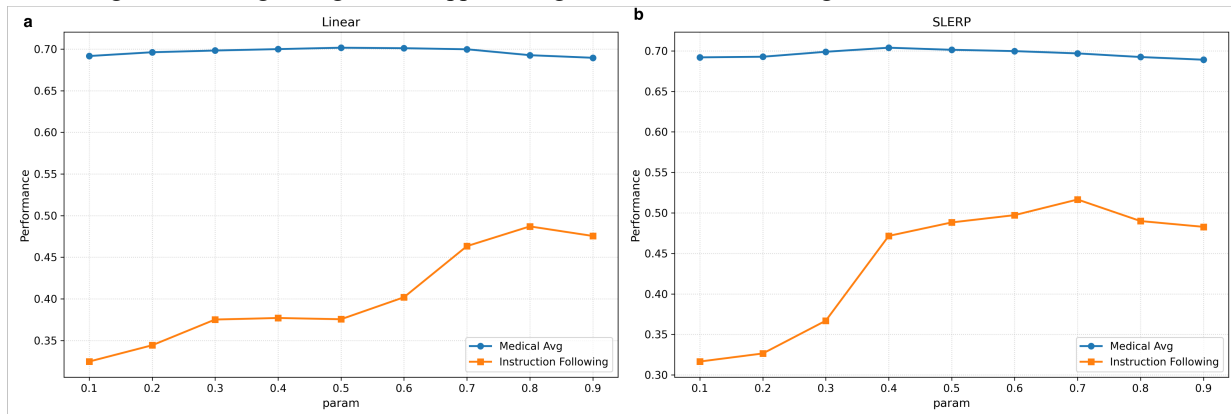

**Figure 3.** Non-monotonic performance trajectories of merged models across varying interpolation weights for (a) Linear and (b) SLERP strategies. The x-axis represents the merge ratio ($\alpha$ or $t$), where a value of 0 corresponds to the clinical foundation model and 1 to the general instruct model. Medical Avg represents the unweighted mean performance across the clinical benchmark suite, and Instruction Following denotes the average of IFEval strict (prompt-level and instance-level) scores.

To assess merged models for full-training scenario, we selected the strongest zero-shot merged model on clinical benchmarks, specifically SLERP ($t = 0.4$) and Linear ($\alpha = 0.5$), for full-parameter supervised fine-tuning (SFT) across five clinical generation tasks. **Table 1.** reports both zero-shot and post-SFT performance for the merged models, together with GatorTronLlama (clinical foundation), GatorTronLlama_SFT (clinical instruction-tuned baseline), and Llama-3.1-8B-Instruct (general instruction baseline).

SFT consistently improved performance across all models, with notable gains in the IU-Xray task (increasing from ~0.18–0.20 to 0.634–0.647). The merged models demonstrate better zero-shot initializations and maintained this advantage post-fine-tuning. The SLERP-merged model attained the highest post-SFT composite scores across all five tasks, consistently outperforming both the clinical foundation model and the instruction-tuned clinical baseline (GatorTronLlama_SFT). For example, SLERP achieved post-SFT scores of 0.2950 on MIMIC-BHC and 0.4099 on ACI-Bench. These results indicate that Pareto-optimized merging provides an effective initialization for supervised clinical adaptation.

**Table 2.** Finetuning performance on clinical downstream generation tasks

| Model | MIMIC-BHC | | ACI-Bench | | MTS-Dialogue | | Problem List | | IU-Xray | |
|---|---|---|---|---|---|---|---|---|---|---|
| | 0-shot | SFT | 0-shot | SFT | 0-shot | SFT | 0-shot | SFT | 0-shot | SFT |
| GatorTronLlama | 0.2271 | 0.2643 | 0.2854 | 0.3685 | 0.2491 | 0.4512 | 0.2084 | 0.3307 | 0.1846 | 0.6347 |
| GatorTronLlama_SFT | **0.2481** | 0.2822 | 0.2969 | 0.3928 | 0.2577 | 0.4550 | 0.2153 | 0.3385 | 0.1950 | 0.6399 |
| Llama-3.1-8B-Instruct | 0.2332 | 0.2708 | 0.2995 | 0.3671 | 0.2660 | 0.4441 | 0.1905 | 0.3305 | 0.1949 | 0.6341 |
| Linear Merge | 0.2397 | 0.2728 | 0.3058 | 0.3883 | 0.2699 | 0.4394 | 0.2117 | 0.3147 | 0.1955 | 0.6347 |
| SLERP Merge | 0.2478 | **0.2950** | **0.3194** | **0.4099** | **0.2754** | **0.4587** | **0.2340** | **0.3561** | **0.2020** | **0.6471** |

Scores are computed as a composite metric, defined as the arithmetic mean of ROUGE-1/2/L, BLEU, METEOR, and BERTScore.

To evaluate adaptation potential under limited supervision, we conducted few-shot fine-tuning on radiology report summarization (IU-Xray) and active problem list generation (**Figure 4**). We evaluated performance using the same composite score defined in Table 2, varying the training budget from 8 to 512 examples. Across both tasks, SLERP merged model demonstrated substantially more sample-efficient learning trajectories than the clinical foundation model. On the IU-Xray dataset, the SLERP checkpoint established an advantage at the lowest supervision level, achieving a score of 0.4778 with 8 examples. This exceeded both GatorTronLlama (0.4502) and GatorTronLlama_SFT (0.4462). This advantage persisted with additional data: SLERP trained on 64 examples (0.5167) matched the performance of the foundation model trained on 256 examples (0.5157). At the maximum 512-example threshold, SLERP maintained its lead (0.5281) over all baselines.

The problem list generation task yielded comparable trends in data efficiency. SLERP improved early-stage learning, scoring 0.2407 at 8 examples compared to 0.2136 for the foundation model. Notably, SLERP achieved a score of 0.2587 with 64 examples, slightly surpassing the foundation model's performance at the full 512-example level (0.2581), and reached the highest overall performance at 512 examples (0.2723). Overall, these findings suggest that model merging substantially reduces the annotated data required to reach baseline clinical performance.

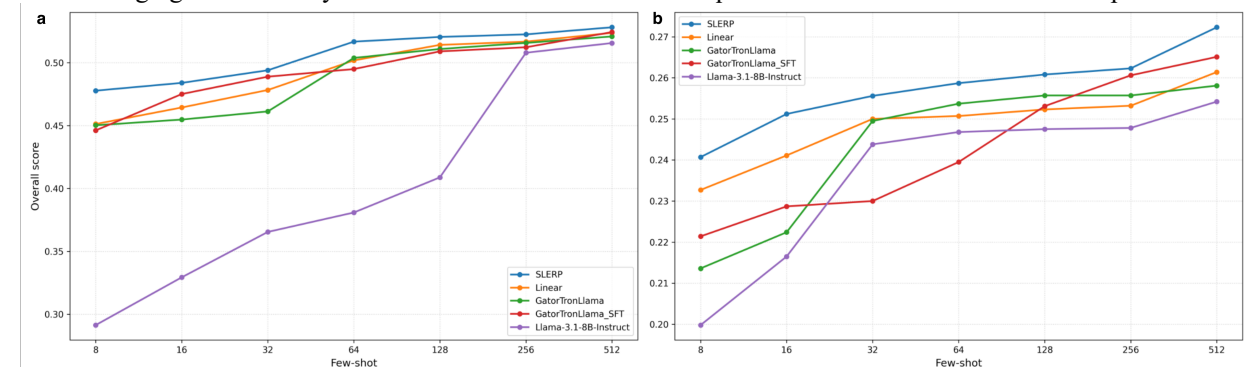

**Figure 4.** Training efficiency of merged models under limited supervision. (a) radiology report summarization and (b) active problem list generation. The x-axis represents the number of labeled examples, and the y-axis represents the composite overall score.

**Discussion**

In this study, we demonstrated that model merging provides a viable, low-resource strategy for adopting open-source general-purpose LLMs to the medical domain. By interpolating the parameters of a domain-specific clinical foundation model (GatorTronLlama) and a general instruct model (Llama-3.1-8B-Instruct) through model merging, we derived merged models with better tradeoffs between clinical tasks and instruction following ability, countering the catastrophic forgetting issue. Compared to instruction fine-tuning or Reinforcement Learning from Human Feedback (RLHF), this approach doesn't require additional parameter optimization and reduces compute cost.

A primary challenge in adapting general-purpose LLMs for clinical tasks through post-training, such as SFT, is the "forgetting" issue, where the model fine-tuned using task-specific clinical datasets often forgets the instruction-following ability derived in the instruction-training phase of the general LLMs. Our zero-shot evaluation suggests that model merging helps mitigate this forgetting issue. Our model merging strategy via Spherical Linear Interpolation (SLERP) could improve the performance on clinical tasks (e.g., a 1.43% improvement over the foundation model at t = 0.4), while reserving 98.3% of the instruction-following ability from the general-purpose instruction-tuned LLMs. The observed non-monotonic relationship between merge ratio and clinical performance indicates a critical "sweet spot". This non-monotonicity suggests that weight interpolation is beyond a simple linear projection of weight; rather, it could transfer the instruction-following ability while keeping a good performance on clinical tasks.

A key contribution of this study is the adaptation of general-purpose LLMs to the clinical domain. Curating large-scale, high-quality instruction datasets is very expensive in the medical domain. Our post-SFT and few-shot experiments provide evidence that model merging can serve as an efficient solution to transfer the instruction following ability of open-source LLMs into medical LLMs. Across five clinical documentation tasks, including radiology report summarization and problem list generation, the SLERP-merged model consistently demonstrated remarkably low training cost. For instance, on the IU-Xray dataset, the SLERP merged model fine-tuned on 64 examples achieved performance comparable to the clinical foundation model trained on 256 examples. Similar data efficiency was observed in active problem list generation, where the 64-example SLERP merged model surpassed the 512-example foundation model. Overall, these results suggest that model merging can reduce the labeled-data requirements needed to reach competitive performance for clinical applications.

This study has limitations. Our evaluation relies primarily on automatic metrics (e.g., ROUGE, BLEU, BERTScore) and established benchmarks. While this provides a reproducible quantitative assessment, clinical text generation ultimately requires strict factual correctness and safety. Future work will conduct rigorous clinician adjudication using structured rubrics to formally evaluate factual accuracy, hallucination rates, and clinical safety. Our search grid was relatively constrained; exploring a broader density and weight grid, or alternative conflict-resolution thresholds, may bring further improvements. Finally, our experiments were conducted at the 8-billion-parameter scale. Future work should investigate whether these merging dynamics hold true for larger-scale models in the clinical domain.

**Acknowledgment**
This study was partially supported by grants from the Patient-Centered Outcomes Research Institute® (PCORI®) Award ME-2023C3-35934, the PARADIGM program awarded by the Advanced Research Projects Agency for Health (ARPA-H), National Institute on Aging U24AG098157, National Institute of Allergy and Infectious Diseases, NIAID R01AI172875, National Heart, Lung, and Blood Institute, R01HL169277, R01HL176844, National Institute on Drug Abuse, NIDA R01DA057886, R01DA063631, and the UF Clinical and Translational Science Institute. The content is solely the responsibility of the authors and does not necessarily represent the official views of the funding institutions.